\documentclass[pageno]{jpaper}
\usepackage[normalem]{ulem}
\pdfoutput=1
\makeatletter                   
\def\mdseries@tt{m}             
\makeatother                    
\usepackage[plain]{fancyref}
\usepackage[frozencache]{minted} 
\usepackage{color}
\usepackage{placeins}
\usepackage{hyperref}           
\hypersetup{
    colorlinks=true,
    linkcolor=blue,
    filecolor=red,
    urlcolor=magenta,
    breaklinks=true,            
}
\usepackage{breakurl}           

\usepackage{xspace}
\usepackage{amsmath}
\usepackage{bcprules}
\def\OPTIONConf{1}%
\usepackage{joshuadunfield}
\usepackage[numbers]{natbib}
\usepackage{enumitem}
\usepackage[nounderscore]{syntax}
\usepackage{longtable}
\usepackage{booktabs}
\usepackage{caption}
\usepackage{subcaption}
\usepackage{tabu}
\usepackage{amssymb}
\usepackage{mathtools}
\usepackage{authblk}
\usepackage{color}
\usepackage{graphicx}
\usepackage{alltt}
\usepackage{lineno}
\usepackage{proof}
\usepackage{verbatim}
\usepackage{xcolor}
\usepackage{tikz}

\newcommand{\nnvm}[0]{NNVM\xspace}
\newcommand{\tvm}[0]{TVM\xspace}
\newcommand{\relay}[0]{Relay\xspace}

\definecolor{sampldark}{RGB}{72, 71, 135}
\definecolor{uwpurp}{RGB}{51, 0, 111}
\definecolor{uwpurplight}{RGB}{94, 0, 204}
\newcommand{\kwd}[1]{\textcolor{uwpurplight}{\texttt{#1}}}

\usepackage{graphics}
\newcommand{\langl}{\begin{picture}(4.5,7)
\put(1.1,2.5){\rotatebox{60}{\line(1,0){5.5}}}
\put(1.1,2.5){\rotatebox{300}{\line(1,0){5.5}}}
\end{picture}}
\newcommand{\rangl}{\begin{picture}(4.5,7)
\put(.9,2.5){\rotatebox{120}{\line(1,0){5.5}}}
\put(.9,2.5){\rotatebox{240}{\line(1,0){5.5}}}
\end{picture}}

\usepackage{todonotes}

\newcommand{\is}[2][]{&::=&#2\ifthenelse{\equal{#1}{}}{}{&\text{(#1)}}}

\newcommand{\inlineAlt}{~\vert~}
\newcommand{\alt}[2][]{\\&&\vert&#2\ifthenelse{\equal{#1}{}}{}{&\text{(#1)}}}

\newcommand{\altSpace}[3]{\\[#1]&&\vert&#3\ifthenelse{\equal{#2}{}}{}{&\text{(#2)}}}
\newcommand{\grammarname}[1]{\textit{#1}}
\newcommand{\grammarword}[1]{\texttt{#1}}
\newcommand{\op}{\grammarname{op}}
\newcommand{\gvar}{\kwd{@}\grammarname{g}}
\newcommand{\type}{\grammarname{$\tau$}}
\newcommand{\expr}{\grammarname{e}}
\newcommand{\lvar}{\kwd{\%}\grammarname{l}}
\newcommand{\graphVar}{\kwd{\%}\grammarword{graph}}
\newcommand{\real}{\grammarname{r}}
\newcommand{\bool}{\grammarname{b}}
\newcommand{\nat}{\grammarname{n}}

\newcommand{\rettype}{\kwd{$\rightarrow$}\ \type}
\newcommand{\seq}[1]{#1\kwd{,}\ \ldots \kwd{,}\ #1}
\newcommand{\relations}{\kwd{where}\ \seq{\type}}
\newcommand{\tyvar}{\grammarname{tv}}
\newcommand{\tyParam}{\grammarword{T}}

\newcommand{\typeanno}{\kwd{:}\, \type}

\newcommand{\maybe}[1]{\texttt{(}#1\texttt{)?}}

\newcommand{\body}[1]{\kwd{\{} #1 \kwd{\}}}
\newcommand{\param}{\grammarword{x}}

\newcommand{\tyargs}{\kwd{$\langl$} \seq{\type} \kwd{$\rangl$}}
\newcommand{\args}{\kwd{(} \seq{\expr} \kwd{)}}
\newcommand{\bnfrule}[2]{\textsf{#1} & #2}
\newcommand{\basetype}{\grammarname{bt}}
\newcommand{\shape}{\grammarname{s}}
\newcommand{\patt}{\grammarname{p}}

\newcommand{\typename}{\grammarname{tn}}

\newcommand{\rFnRule}{\begin{split}
  \kwd{fn}\ &\maybe{\rTyParamsRule}\\&\rParamsRule\ \maybe{\rettype}\\&\body{\expr}
\end{split}}

\newcommand{\rTupRule}{\kwd{(} \seq{\expr} \kwd{)}}
\newcommand{\rTupProjRule}{e \kwd{.} \nat}
\newcommand{\rIfElse}[3]{\kwd{if}\ \kwd{(} #1 \kwd{)}\ \body{#2}\ \kwd{else}\
\body{#3}}
\newcommand{\rLetRule}{\kwd{let}\ \lvar\ \maybe{\typeanno}\ \kwd{=}\ \expr \kwd{;}\ \expr}
\newcommand{\rAnonLetRule}{\expr \kwd{;}\ \expr}
\newcommand{\rGraphLetRule}{\graphVar\ \kwd{=}\ \expr \kwd{;}\ \expr}
\newcommand{\rLet}[4]{\kwd{let}\ #1\ifthenelse{\equal{#2}{}}{}{\ \kwd{:}\, #2}\ \kwd{=}\ #3 \kwd{;}\ifthenelse{\equal{#4}{}}{}{\ #4}}
\newcommand{\rTyParamsRule}{\kwd{$\langl$} \seq{\tyParam} \kwd{$\rangl$}}
\newcommand{\rParamsRule}{\kwd{(} \seq{\param} \kwd{)}}

\newcommand{\rMatchRule}{\begin{aligned}
  &\kwd{match}\ \kwd{(} \expr \kwd{)}\ \kwd{\{}\\
  &\quad\kwd{|}\ \patt\ \kwd{$\rightarrow$}\ \expr \\
  &\quad \vdots\\
  &\quad\kwd{|}\ \patt\ \kwd{$\rightarrow$}\ \expr \\
  &\kwd{\}}
\end{aligned}}

\makeatletter

\newcommand{\intercCheckNext}[2]{\@ifnextchar\bgroup{\intercGobbleNext{#1}{#2}}{}}
\newcommand{\intercGobbleNext}[3]{#1 #3\@ifnextchar\bgroup{\intercGobbleNext{#1}{#2}}{#2}}
\makeatother

\newcommand{\affiliation}[1]{//\texttt{#1}}
\newcommand{\email}[1]{//\texttt{#1}}

\begin{document}

\title{\relay: A High-Level Compiler for Deep Learning}

\author{Jared Roesch}


\author{Steven Lyubomirsky}

\author{Marisa Kirisame}

\author{Logan Weber}

\author{Josh Pollock}

\author{Luis Vega}

\author{Ziheng Jiang}

\author{Tianqi Chen}

\author{Thierry Moreau}

\author{Zachary Tatlock}

\renewcommand\Authfont{\fontsize{12}{14.4}\selectfont}
\renewcommand\Affilfont{\fontsize{9}{10.8}\itshape}

\affil{
  \texttt{\string{jroesch, sslyu, jerry96, weberlo, joshpol, vegaluis, ziheng, tqchen, moreau, ztatlock\string}@cs.uw.edu} \protect\\
  Paul G. Allen School of Computer Science \& Engineering \protect\\
  University of Washington
}
\date{}
\maketitle

\thispagestyle{empty}

\newcommand{\TQC}[1]{\textcolor{red}{TQ: {#1}}}

\begin{abstract}
Frameworks for writing, compiling, and optimizing
  deep learning (DL) models have recently
  enabled progress in areas like computer
  vision and natural language processing.
Extending these frameworks to accommodate
  the rapidly diversifying landscape of
  DL models and hardware platforms presents
  challenging tradeoffs between
  expressivity, composability, and portability.
We present \relay,
  a new compiler framework for DL.
\relay's functional, statically typed intermediate representation (IR)
  unifies and generalizes existing DL IRs
  to express state-of-the-art models.
The introduction of \relay's expressive IR requires
  careful design of domain-specific optimizations,
  addressed via \relay's extension mechanisms.
Using these extension mechanisms,
  \relay supports a unified compiler that
  can target a variety of hardware platforms.
Our evaluation demonstrates \relay's competitive performance for a
  broad class of models and devices
  (CPUs, GPUs, and emerging accelerators).
\relay's design demonstrates how a unified IR can provide
  expressivity, composability, and portability
  without compromising performance.

\end{abstract}

\section{Introduction}
\label{sec:intro}

Deep learning (DL) has radically transformed domains like
  computer vision and
  natural language processing (NLP)~\cite{yolo, recent_trends_in_nlp}.
Inspired by these successes,
  researchers and companies are continually
  experimenting with increasingly sophisticated DL models and
  developing specialized hardware backends.
DL frameworks for writing, optimizing, and compiling DL models
  reduce the complexity of these tasks,
  which in turn accelerates DL research and product development.

Popular DL compiler intermediate representations (IRs) offer different tradeoffs
  between expressivity, composability, and portability~\cite{
    tensorflow, pytorch_ad, chainer_learningsys2015, tangent, theano, glow}.
Early frameworks adopted IRs
  specialized for then-state-of-the-art models and/or
  emerging hardware accelerators.
As a result, non-trivial extensions require
  patching or even forking frameworks~\cite{
    tf_fold, tf_lite, tangent, tf_eager, xla, glow, torchscript}.
Such \textit{ad hoc} extensions can improve expressivity
  while maintaining backwards compatibility with existing execution mechanisms.
However, they are difficult to design, reason about, and implement,
  often resulting in modifications that are mutually incompatible.

Let us consider a hypothetical scenario that exemplifies
  IR design tensions in DL compilers.
Suppose a machine learning engineer wants to write
  an Android app that uses sentiment analysis to
  determine the moods of its users.
To maintain privacy, the app must run completely on-device,
  i.e., no work can be offloaded to the cloud.
The engineer decides to use a variant of TreeLSTM,
  a deep learning model that uses a tree structure~\cite{tree_lstm}.
Unfortunately, current frameworks' IRs cannot directly encode trees,
  so she must use a framework extension
  like TensorFlow Fold~\cite{tensorflowfold}.

Suppose that after adapting the model to run on her phone,
  the out-of-the-box performance of her
  model on her particular platform is not satisfactory, requiring her to optimize it.
She chooses to employ \textit{quantization}, an optimization that
  potentially trades accuracy for performance by replacing
  floating-point datatypes with low-precision ones.
Although researchers have developed a variety of quantization
  strategies, each of which makes use of different bit-widths, rounding
  modes, and datatypes, our engineer must use a strategy supported
  by existing frameworks~\cite{gustafson2015end, tf_lite_ops_compat, glow_quant}.
Unfortunately, frameworks only provide support for a small number
  of strategies, and supporting new quantization strategies is non-trivial.
Each combination of operator, datatype, bit-width, and
  platform requires unique operator implementations.
Optimizations like operator fusion exacerbate this combinatorial explosion,
  further increasing
  the number of unique implementations required.
Furthermore, if a framework doesn't have specific support for
  the target phone model she cannot take advantage of specialized deep learning
  instructions or coprocessors~\cite{apple_neural_engine}.

The scenario above highlights the three-pronged \textit{extensibility challenge}
  for DL IRs:
\begin{enumerate}[label=\arabic*.]
  \item \textit{Expressivity}: It should be straightforward to write models involving
    control flow, first-class functions and data structures (e.g., trees, graphs, and lists).
  \item \textit{Composability}: It should be straightforward to add and compose new optimizations
    with existing ones (e.g., quantization, operator fusion, and partial evaluation).
  \item \textit{Portability}: It should be straightforward to add new hardware targets
    (e.g., TPU, Inferentia)~\cite{tpuv1, inferentia}.
\end{enumerate}

Previous IRs have struggled to address these challenges, treating each
  component of the framework as a disconnected set of programming tasks.
Operators are defined in low-level languages like C++,
  connected by a dataflow graph, and then scripted
  in a host language like Python.
Consequently,
  program analyses cannot cross language boundaries between components,
  inhibiting optimization and deployment.
Learning from previous IRs, we have designed \relay,
  which features a principled approach to addressing extensibility
  and improves expressivity, composability, and portability
  over previous frameworks.
We make the following contributions:
\begin{itemize}
  \item The \relay IR, a tensor-oriented, statically typed
    functional IR,
    which we describe in Section \ref{sec:design}.
  \relay's design is motivated by the insight that functional IRs, used by
  languages from the ML family\footnote{``ML'' as in ``Meta Language,'' not
  ``Machine Learning''} can be readily adapted to support DL.
  With its \textit{expressive} semantics,
    including control flow, data structures, and first-class functions,
    \relay can represent entire state-of-the-art models.
  \item The insight that common features in ML frameworks,
    such as quantization and shape inference,
    can be reframed as standard compiler passes.
  By using this reframing we can tap into
    decades of traditional compilers research to design
    \textit{composable} optimization passes.
  \item
    A platform-agnostic representation of operators and domain specific
      optimizations which work in concert to provide \textit{portability}
      across hardware backends.
\end{itemize}

We evaluate \relay on several systems and over a diverse set of vision and NLP workloads to
  demonstrate that (1) \relay enables \emph{expressive} programs via a large breadth
  of models, (2) \relay supports \emph{composition} of program-level optimizations
  such as quantization and fusion, and (3) \relay provides
  \emph{portability} by targeting a number of hardware backends.
Not only does \relay provide these three properties, we do so while also demonstrating
  competitive performance.
\relay is an open-source academic project.\footnote{\relay is publicly available at [redacted for review].}
  It has been deployed at a popular web service provider,
    a telecommunications and consumer electronics manufacturer,
    and a social media company, among others.

\section{Related Work}
\label{sec:related}


The acceleration of deep learning is an active topic of research and is
  cross-disciplinary by nature.
The dominant platforms for deep learning are TensorFlow, PyTorch, and MxNet.
Research on these frameworks cuts across all abstraction levels and
  involves experts from machine learning, systems, architecture, and programming languages (PL).
We first discuss the evolution of modern DL frameworks,
  then the lower-level components DL frameworks have incorporated to gain performance
  (i.e., low-level tensor compilers and DL compilers),
  and finally, we turn to approaches from the PL community.

\subsection{Deep Learning Frameworks}

In the early days of deep learning, practitioners and researchers would program
  in general-purpose languages like Python utilizing
  scientific computing libraries like NumPy,
  which provide low-level \textit{operators} such as matrix multiplication.
In order to accelerate model execution,
    frameworks supporting accelerators such as GPU were introduced~\cite{theano} .
Early frameworks represented models as directed ``computation graphs'',
    where each node represents an operator,
    and each edge represents the flow of data from one operator to another.
Computation graphs provide a limited programming model,
    enabling straightforward mapping of operators onto GPUs.
Large technology companies,
    such as Google, Facebook, and Amazon,
    drive the development of frameworks,
    and consequently,
    each company has its own stack consisting
    of the core framework (TensorFlow~\cite{tensorflow}, PyTorch~\cite{pytorch}, MxNet~\cite{mxnet}),
    compilers(XLA~\cite{xla}, Glow~\cite{glow}, TVM~\cite{tvm_osdi18}),
    and hardware accelerators (TPU~\cite{tpuv1}, GraphCore, Inferentia~\cite{inferentia}).
Frameworks can be roughly categorized into those which support \textit{static} computation graphs
  and those which support \textit{dynamic} computation graphs.
Frameworks which use static graphs are said to be \textit{define-and-run} frameworks,
  whereas frameworks which use dynamic graphs are said to be \textit{define-by-run} frameworks.

\subsubsection*{Define-And-Run Frameworks}

TensorFlow, Caffe~\cite{caffe}, and Theano~\cite{theano} are define-and-run frameworks.
Static graphs represent a whole-program,
  enabling optimization and simplified deployment,
  by removing the need for a host language like Python.
TensorFlow (TF) extends pure dataflow graphs with \textit{control edges}
      to emulate the functionality of \verb|if| and \verb|while|.
TF's representation captures many state-of-the-art models,
      provides support for heterogeneous hardware back-ends,
      and enables reverse-mode automatic differentiation~{\cite{ad_survey, tensorflow}}.
TF's encoding of control has limitations, as control-flow structures
    do not clearly map to familiar control-structures, instead using specialized
    encodings which make adapting traditional optimizations challenging.
Furthermore,
    unmodified TensorFlow does not support building models where the shape of
    the computation graph is dependent on the input,
    frustrating researchers who wish to experiment with complex models.
TensorFlow Fold addresses this \textit{particular} limitation~\cite{tensorflowfold}
    but offers no general and extensible solution.
The crux of the problem is the lack of generic mechanisms for users to
    define new control flow combinators (e.g., \verb|fold|) and data types.

\subsubsection*{Define-By-Run Frameworks}
PyTorch~\cite{pytorch_ad}, Gluon~\cite{gluon}, Chainer~\cite{chainer_learningsys2015},
    and TensorFlow eager-mode~\cite{tf_eager} are define-by-run frameworks which
    attempt to address the challenges of previous work.
The approach popularized by PyTorch is to use a host language (e.g., Python)
    to eagerly execute operations while simultaneously building a computation graph
    as a side effect.
By using the full host language,
  its features may be used to provide a highly expressive programming model to users.
However, dynamic frameworks construct a graph \textit{per program trace} and must re-optimize when
    the graph topology changes, costing CPU cycles and incurring communication overhead between the host
    machine and accelerators.
Instead of just representing traces, \relay combines the advantages of both worlds by
    representing the whole program ahead of time,
    while supporting constructs like control flow, first-class functions, and data structures.

\subsection{Low-Level Tensor Compilers}
Low-level tensor compilers are focused on the production
    of high-performance operators which implement compute-intensive
    operations such as matrix multiplication or convolution.
There are a number of competing approaches,
    both from academic and commercial entities, such as
    TVM~\cite{tvm_osdi18}, Halide~\cite{halide}, Tensor Comprehensions(TC)~\cite{tensor_comprehensions},
    and Diesel~\cite{diesel}.
The most notable designs are either inspired by the
    compute-schedule split introduced by Halide
    and adapted by TVM, or the polyhedral framework,
    as used by TC and Diesel.
Operator compilers perform code generation for sets of scalar loop nests,
    but only represent a restricted subset of a whole program, ignoring details such as
    memory allocation/management, data structures, closures, and arbitrary control flow.
\relay focuses on composing generic operators, and the surrounding program
    into an efficiently orchestrated DL program.

\subsection{Deep Learning Compilers}

DL frameworks have adopted compilers
    to tackle both performance and portability
    for existing applications, most notably
    XLA~\cite{xla}, Glow~\cite{glow}, nGraph~\cite{ngraph}, ONNC~\cite{onnc},
    PlaidML~\cite{plaidml}, and ModelCompiler.
These \textit{graph compilers} use computation graph IRs and provide
    lowering onto a variety of targets.
Often graph compilers only perform high-level optimizations
    and then offload to vendor-specific libraries.

Due to their limited programming model, they
    provide the same functionality as \relay with
    a more limited language.
The most comparable points to \relay are recent
    developments in the TensorFlow and PyTorch
    ecosystems of MLIR and TorchScript, respectively.
Google introduced MLIR as a path forward for
    unifying its myriad of IRs.
Upon first examination MLIR might appear to be
    a replacement for XLA and related TF compiler
    efforts, but it is not that.
MLIR is shared infrastructure for constructing
    a set of interoperating IR ``dialects'' which
    can be used to construct compilers.
The MLIR project is working on IR dialects
    for TF's IR and a low-level polyhedral IR,
    but does not yet have an end-to-end solution for
    deep learning built upon MLIR, the insights in
    this paper can guide MLIR's dialect development.

TorchScript is a high-level Python-like IR developed as the first
    layer of PyTorch's JIT compiler.
PyTorch (since v1.0) can rewrite a subset of user programs into
    TorchScript, an idealized subset of Python.
TorchScript can then be executed by the TorchScript VM or JIT-compiled to a target platform.
TorchScript sits many layers above code generation and must accommodate
    the flexible semantics of Python, which rules out entire classes of static analysis.
In order to optimize away this dynamic behavior, TorchScript has
    a profiling JIT mode which identifies stable program traces
    during execution.
These stable static traces can then be optimized by lower-level
    compilers such as Glow or \relay to perform the last level of code generation.
Microsoft released ModelCompiler, a system for efficiently compiling RNNs defined
    in CNTK to CPU.
ModelCompiler uses Halide to represent low-level operations, but lacks
    the expressivity of the \relay IR and only demonstrates support for CPUs.

\subsection{Programming Languages for Deep Learning}
\label{sec:pl_techniques_in_dl}

In recent years, the design of new programming languages,
    or the augmentation of existing ones, has become
    a popular area of research.
New languages designed for machine learning and related
    tasks include Lantern~\cite{lantern}, Lift~\cite{lift_lang}, Flux.jl~\cite{fluxjl}
    AutoGraph~\cite{moldovan2018autograph}, Swift for TensorFlow~\cite{tf_swift},
    and JAX~\cite{jax}.
Lantern \cite{lantern} is the most related work to \relay as it can
    be used as a code generator.
Lantern is a deep learning DSL in Scala
    that uses lightweight modular staging (LMS) to lower code into C++ and CUDA.
Lantern's defining feature is the use of delimited continuations to perform
    automatic differentiation.
Delimited continuations provide an elegant algorithm for AD,
    only requiring local transforms, but incurs cost of
    heap allocated structures, and a less straightforward
    mapping to define-by-run frameworks.
Lantern solves this problem by using a CPS transform which
    complicated further optimization and code generation.
Lantern does not yet support hardware accelerators, and
    does not focus on full program optimizations.
The alternative approach is the augmentation of languages to support deep learning,
  the most notable being systems like AutoGraph, Flux.jl, Swift for TensorFlow,
  and JAX.
These systems are designed to be user-facing programming
    environments for deep learning and use a compiler IR
    to generate code.
For all intents and purposes \relay could be the IR in
    question, therefore  \relay complements these systems well by
    providing a more expressive IR to map computation onto.

\section{Design}
\label{sec:design}

\relay's expressive high-level IR is designed to support
  complex models while abstracting over hardware-specific
  implementation details to enable hardware agnostic program
  analysis and optimization.
Rather than invent an entirely new language,
  \relay's IR design is based on IRs used by the well-studied ML family of
  functional programming languages (e.g., SML and OCaml).
These IRs are expressive enough to capture general-purpose programs
  (including control flow, first-class functions, and data types)
  and have clearly specified semantics (e.g., lexical scope and controlled effects).
By borrowing from PL literature,
  we can apply program analysis and optimization techniques from decades of research~\cite{haskell_vector}.

\relay's IR takes a small functional core and enriches it with domain-specific additions---namely,
  the inclusion of tensors and operators as expressions
  and a novel tensor type system design to support tensor shapes.
Our principled design
  enables the import of existing models from deep learning frameworks and exchange formats,
  the implementation of a number of domain-specific optimizations,
  and efficient deployment across a variety of targets.
In the remainder of this section,
  we describe the IR design in further detail
  and explore the ramifications of this design on the compilation stack.

\subsection{IR}
  \begin{figure}[t]
    \begin{jmpgrammar}
      \bnfrule{Expr}{\expr} \is[local var]{\lvar}
        \alt[global variable]{\gvar}
        \alt[constant tensor]{\kwd{const} \kwd{(} \texttt{(}\real\inlineAlt\bool\texttt{)} \kwd{,} \shape \kwd{,} \basetype \kwd{)}}
        \alt[call]{\expr \maybe{\tyargs} \args\vspace{0.2em}}
        \alt[let]{\rLetRule}
        \alt[\kwd{let}\ \kwd{\%}\_\ \kwd{=}\ \expr\kwd{;}\ \expr]{\rAnonLetRule}
        \alt[graph let]{\rGraphLetRule}
        \altSpace{0.5em}{function}{\rFnRule}
        \altSpace{1em}{tuple formation}{\rTupRule}
        \alt[tuple proj.]{\rTupProjRule}
        \alt[if-else]{\rIfElse{\expr}{\expr}{\expr}}
        \altSpace{0.5em}{pattern match}{\rMatchRule}
        \altSpace{1em}{operator}{\op}
        \alt[new ref]{\kwd{ref}\kwd{(}\expr\kwd{)}}
        \alt[get ref]{\kwd{!} \expr}
        \alt[set ref]{\expr \kwd{:=} \expr}\\\\
      \bnfrule{Type}{\type} \is[base type]{\basetype}
        \alt[shape]{\shape}
        \alt[tensor type]{\kwd{Tensor} \kwd{[} \shape \kwd{,} \basetype \kwd{]}}
        \alt[type variable]{\tyvar}
        \alt[function type]{
          \begin{split}
          \kwd{fn}\ &\rTyParamsRule\\
          &\kwd{(} \seq{\type} \kwd{)}\ \rettype\\
          &\maybe{\relations}
          \end{split}
          }
        \alt[ref type]{\kwd{Ref} \kwd{[} \type \kwd{]}}
        \alt[tuple type]{\kwd{(} \seq{\type} \kwd{)}}
        \alt[type call]{\type \kwd{[} \seq{\type} \kwd{]}}
        \alt[type name]{\typename}
    \end{jmpgrammar}
    \caption{\textmd{The BNF Grammar for the \relay{} language.}}
    \label{fig:short_bnf}
  \end{figure}

The \relay IR is designed
  to subsume the functionality of computation graph-based IRs
  while providing greater faculties for abstraction and control flow.
We present \relay's design by incrementally building up to the full IR
  starting from a subset that corresponds to a simple computation graph.
Deep learning models fundamentally operate on tensors.
Hence, \relay's primary value type is a tensor and operators are included as language primitives
  (see the \verb|tensor constant| and \verb|operator| rules in Figure \ref{fig:short_bnf}).
\relay leaves the implementation of each operator opaque; the operators
  are represented by a lower-level IR, which is optimized independently.
A computation graph, in its simplest form, is a directed acyclic
  graph with multiple inputs and a single output.
\relay uses three constructs to support these simple graphs:
  (1) \verb|variable|, (2) function \verb|call|,
  and (3) \verb|operator|; see Figure~\ref{fig:short_bnf} for the corresponding rules.

\subsubsection*{Multiple Outputs}

Computation graph IRs have primitive support for multiple outputs
  because many tensor operators require it.
For example, the \verb|split| operator separates a tensor along a given axis
  and returns each component.
In \relay, multiple outputs can be modeled as tuples,
  requiring only two rules: \verb|tuple formation| and \verb|tuple projection|.

\subsubsection*{Let}

By construction, computation graphs enjoy implicit sharing of subcomputations
  via multiple outgoing dependency edges.
Implicit sharing is often implemented via pointers that uniquely identify subgraphs,
  a property useful for both execution and analysis.
Previous frameworks often obtain this sharing by using a host
  language's name binding to construct a graph (e.g., by binding a Python variable
  to a subgraph and using that variable to construct other subgraphs).
General-purpose programming languages, on the other hand, provide \textit{explicit}
  sharing via binding constructs, such as \verb|let|.
In programs free of scope, ordering, and effects, implicit sharing
  and explicit sharing are semantically equivalent.
However, in practice, user programs rely on effects and ordering,
  requiring previous approaches to provide workarounds.
For example, TensorFlow's Eager Mode inserts dummy control edges
  in its generated graphs to impose effect ordering.
The lack of lexical scope in computation graphs complicates language features,
  like first-class functions and control flow,
  and reduces the precision of traditional analyses,
  such as liveness,
  because the high-level program structure is absent~\cite{funarg, funarg_sol}.
The addition of a humble \verb|let| binding, a central concept in functional languages,
  provides explicit sharing and a solution to the problems outlined above.

\subsubsection*{Control Flow}

\begin{figure*}[htb!]
  \begin{tabular}{ccc}
  \begin{minipage}{0.4\textwidth}
  \begin{minted}[fontsize=\small]{python}
i = tf.constant(1)
j = tf.constant(1)
k = tf.constant(5)

def c(i, j, k):
  return
    tf.equal(
      tf.not_equal(
        tf.less(i + j, 10),
        tf.less(j * k, 100)),
       tf.greater_equal(k, i + j))
def b(i, j, k): return [i+j, j+k, k+1]
tf.while_loop(c, b, loop_vars=[i, j, k])
  \end{minted}
  \end{minipage}
& \hspace{-2.0em}
\begin{Huge}
  $\Rightarrow$
\end{Huge}
&
  \begin{minipage}{0.5\textwidth}
  \begin{minted}[fontsize=\footnotesize]{python}
  fn %while_loop(
    %lvar0: Tensor[(1,), int32], %lvar1: Tensor[(1,), int32],
    %lvar2: Tensor[(1,), int32]) {
    %0 = add(%lvar0, %lvar1)
    %1 = less(%0, meta[Constant][0])
    %2 = multiply(%lvar1, %lvar2)
    %3 = less(%2, meta[Constant][1])
    %4 = not_equal(%1, %3)
    %5 = add(%lvar0, %lvar1)
    %6 = greater_equal(%lvar2, %5)
    if (min(equal(%4, %6))) {
      %9 = add(%lvar0, %lvar1)
      %10 = add(%lvar1, %lvar2)
      %11 = add(%lvar2, meta[Constant][2])
      %while_loop(%9, %10, %11)
    } else { (%lvar0, %lvar1, %lvar2)
    }
  }
  %while_loop(meta[Constant][3], meta[Constant][4], meta[Constant][5])
  \end{minted}
  \end{minipage}
  \end{tabular}
  \caption{\textmd{
    A simple TensorFlow loop in the user-facing DSL and the \relay
      loop produced by automatically converting it.
    Note the TensorFlow while loop corresponds neatly to a tail recursive
      function.
    The \relay text format supports a ``metadata'' section which functions
      as a constant pool among other things.
    \texttt{meta[Constant][n]} represents the \texttt{n}-th constant in the
      pool.
  }}
  \label{fig:tf_to_relay_loop}
  \end{figure*}

Emerging models, particularly in the domain of natural language processing, increasingly
  rely on data-dependent control flow, forcing frameworks based on computation graph IRs
  to incorporate control flow, often through \textit{ad hoc} and difficult-to-extend constructs.
For example, TensorFlow Fold~\cite{tf_fold} extends TF with special combinators that
  dynamically compute a graph for each shape permutation;
  these high-level constructs are opaque to further optimizations.
The functional programming community has demonstrated that recursion and pattern matching are sufficient
  to implement arbitrary combinators for control flow and iteration (e.g., maps, folds, and scans).
To support the definition of functional combinators
  we enrich \relay with two more language
  features to implement arbitrary combinators: \verb|if| and first-class recursive functions.

\subsubsection*{First-Class Functions}

A computation graph is a single computation
  from multiple inputs to multiple outputs.
While it is tempting to reinterpret a graph as a function,
  graphs lack functional abstraction and named recursion.
The addition of first-class named functions dramatically increases
  \relay's expressivity, allowing it to encode generic
  higher-order functions and thus capture higher-level program structure.
First-class functions also enable simpler implementations
  of importers that map higher-level programs to our IR.
For example, an instance of TensorFlow's looping construct \verb|tf.while_loop|
  can be represented as a single specialized loop function
  or a generic fold over the loop state.
See Figure~\ref{fig:tf_to_relay_loop} for an example of this conversion (via
  the \relay TensorFlow frontend).

\subsubsection*{Data Abstraction}
Many models make use of additional data types beyond
  tuples, such as lists, trees, and graphs~\cite{char-rnn, tree_lstm, graph_lstm}.
\relay borrows from functional languages
  a generic and principled method of extension:
  algebraic data types (ADTs).
To support them, we add mechanisms for
  (1) type declaration and
  (2) pattern matching.
This final addition results in a strict functional language,
  closely resembling the core of languages like OCaml and SML.
The increase in expressivity introduced by the \relay IR introduces
  new optimizations challenges, which we
  discuss in Sec.~\ref{sec:optimizations}.

\subsection{Type System}
\label{subsec:type_system}

\relay's type system is essential
  to optimizations.
Typing guarantees both well-formedness of the program
  and provides crucial tensor shape information to perform allocation,
  check correctness, and facilitate loop optimizations.
Shape information is also valuable for data layout transformations and tensorization,
  two transformations often demanded by hardware accelerators.
In computation graph IRs, only numeric data types
  and shapes are tracked for each operator.
Symbolic shapes (i.e., shape polymorphism) are only handled
  dynamically, inhibiting certain types of optimizations.

It is possible to model arbitrarily complex static properties, such
  as shape information, with a dependent type theory~\cite{selsam_certigrad}, but such
  a design incurs significant user complexity.
By incorporating shape analysis into a broader type system,
  \relay's type system balances the desire for static tensor shapes
  with usability.
In this subsection, we describe how to extend a polymorphic type system with shape
  information and type inference with shape inference.

\subsubsection*{Tensor Types}

The primitive value in \relay is a tensor, which has
  a shape and a base type (\verb|tensor type| in Figure \ref{fig:short_bnf}).
Base types describe the elements of tensors by tracking
  the bit width,
  the number of lanes (for utilizing vectorized intrinsics),
  and whether the type is floating point or integral.
To ensure \relay can offload tensor computation to devices
  with greatly varying architectures,
  \relay tensors may only contain base types,
  preventing, for example, tensors of closures.
The shape of a tensor is a tuple of integers describing the tensor's dimensions.
A dimension may be a variable or arithmetic expression that indicates how the
  output shape of an operator depends on those of its inputs.
Functions may be polymorphic over shapes, which results
  in shape constraints that must be solved during type inference.
Sec.~\ref{sec:inference} describes the process.
\relay also supports a special shape called \verb|Any|, which is used
  to mark a dynamic shape when static relationships are not profitable
  to model.

\subsubsection*{Operators and Type Relations}
Operators are one of the key primitives that differs from those of
  general-purpose programming languages.
\relay's use of opaque operators enables backends to choose different
  lowering strategies based on the hardware target.
\relay's operator set is extensible, meaning that users may add new operations.
Supporting common or user-defined tensor operators requires a type system that can
  adapt to complex shape relationships between input and output types
  (e.g., elementwise operators with broadcasting semantics).

To handle the constraints between operators' argument shapes, \relay's type system
  introduces type relations.
A type relation is implemented as a function in the
  meta-language and represents a symbolic relationship between
  the input and output types.
When developers add a new operator to \relay, they may constrain its
  type with an existing relation or add their own.
Function types may include
  one or more type relations over a subset of the argument types and the return type.
The type checker enforces that these relationships hold at each call site.

\subsubsection*{Type Inference}
\label{sec:inference}

To incorporate type relations into \relay's type system, we enrich
  a Hindley-Milner-style type inference algorithm with
  a constraint solver.
\relay's inference algorithm has three steps: first, it
  performs a pass over the AST, generating types and a set of relations,
  then it solves the incurred constraints,
  and finally annotates each sub-expression with its inferred type.


When the type inference algorithm visits a function call site, the function's type relations are
  instantiated with the concrete argument types at the call site.
Each instantiated relation is added to the queue of relations to solve.
The relationship between a call's type variables and relations is added as an edge to
  a bipartite dependency graph where the two disjoint sets are type variables and type relations.
Traditional unification constraints are represented using a modified union-find structure that
  integrates with this dependency graph.

Once the queue is populated, the algorithm will dequeue a relation and attempt to solve it.
There are two cases when solving a type relation:
\begin{enumerate}
  \item If all the relation's type variables
  are concrete, we the relation function. If that function returns true, the
  constraint is discharged. Otherwise, type checking fails.
  \item If any type is fully or partially symbolic, the
    algorithm will propagate
    existing concrete type information via unification.
  All relations affected by new assignments to type
    variables (as determined by the dependency graph)
    are moved to the beginning of the queue.
  If the current type relation is now completely solved, we
  discard it to avoid unnecessarily visiting it again.
\end{enumerate}

We run this to fixpoint or until the queue is empty.
If the queue is non-empty and no progress is made between iterations,
  then at least one variable is underconstrained and inference fails.
Note that a type relation's implementation can
  compromise type soundness, as they are axiomatic descriptions
  of operations implemented outside of \relay.
In practice, the number of type relations needed to express \relay's
  operators is small, and their implementations are straightforward
  and amenable to exhaustive testing.

\subsection{Compiler Framework}

The process for compiling \relay proceeds in three stages.
First, the frontend converts input formats into the \relay IR.
Next, the \relay compiler typechecks and optimizes the program
  to produce the final program.
After performing optimizations,
  the \relay backend transforms
  the \relay program into a form that can be executed on
  the intended hardware, based on the specified execution mechanism.
The backend additionally lowers \relay operators into a TVM expression,
  computes a schedule for the final TVM expression, and lowers it into
  native code.

\subsubsection*{Frontend}

There are several ways to write an \relay program.
A user can build an in-memory representation of
  a program in C++ or Python,
  parse one written in the \relay text format,
  load one from the on-disk serialization format,
  or import one from popular frameworks and interchange formats
    (e.g., TensorFlow, MxNet, Keras, DarkNet, and ONNX).
Many frameworks and interchange formats use static computation graph-based representations,
  which can easily be translated into \relay.
A greater challenge is translating frameworks
  with a richer computation model such as TensorFlow (TF).
TF supports control flow and includes \verb|TensorArray|, a write-once
  tensor container.
We can extract the loop structure out of the TF graph, converting
  it to an \relay loop, and transform the \verb|TensorArray| into an \relay list.
Once new deep learning languages and IRs under development
  are stable it is likely they can be translated into \relay (see
  Section~\ref{sec:pl_techniques_in_dl}).
PyTorch provides an expressive programming model, and is a good fit
  for \relay, which has integration into PyTorch's JIT infrastructure,
  enabling users to transparently use \relay for improved performance.

\subsubsection*{Compiler}
Once an \relay abstract syntax tree (AST) is produced,
  the program is optimized by applying a series of \relay-to-\relay
  passes.
Between each pass, \relay performs type inference and checking,
  rejecting malformed programs as well as populating shape and type
  information that passes can utilize.
The \relay compiler supports traditional optimizations
  (e.g., constant folding, common subexpression elimination, and dead code elimination)
  and domain-specific optimizations
  (see Sec.~\ref{sec:optimizations}).

\subsubsection*{Backends}

\relay produces machine-specific code
  by decomposing the problem of code generation into multiple distinct phases.
\relay translates all operators into \tvm expressions
  to produce dense linear algebra kernels~\cite{tvm_osdi18, tensor_comprehensions, halide}.
\tvm produces low-level operators that expect a fixed calling convention,
  as well as preallocated inputs and outputs.
The result is an object file containing hardware-specific implementations of all
  operations.
The remaining \relay program then is executed or compiled,
  with operator invocations replaced by calls to the optimized operators.
By representing operators as \tvm expressions, we can programmatically
  transform them and automatically generate new implementations for the transformed operators.
Optimizations like fusion and quantization
  rely on this novel behavior.
After primitive operators are lowered,
  the remaining \relay program ties
  together operator invocations, allocation, control-flow,
  recursion, and high-level data structures.
There are multiple options for executing the combined full program:
  the \relay interpreter (with JIT compilation),
  an \relay virtual machine,
  the \tvm graph runtime,
  and an experimental \relay ahead-of-time compiler
  that converts programs to C++ to produce a target-specific binary.

\section{Optimizations}
\label{sec:optimizations}


A high-level IR by itself does not provide a path to high-performance code.
Obtaining high-performance models requires domain-specific
  optimizations tailored to deep learning.
In this section, we showcase the use of the \relay compiler framework
  to write general, domain-specific, and target-specific optimizations,
  enabling generation of high-performance code.

\subsection{Operator Fusion}
\label{sec:fusion}

Operator fusion is an indispensable optimization in deep learning compilers.
Fusion enables better sharing of computation, removal of
  intermediate allocations, and facilitates further optimization by
  combining loop nests.
Fusion is known to be the most critical optimization in machine
  learning compilers, but existing fusion techniques
  are closed (working over a fixed set of ops)
  and target-dependent.
Traditional operator fusion algorithms resemble instruction
  selection:
A sequence of operators eligible
  for fusion is first identified and then replaced with a corresponding
  handwritten fused implementation, usually from a vendor-provided library.
For example, if a fused implementation for a GPU operator does not exist in CuDNN,
  it will remain unfused.
More advanced strategies, implemented in XLA, detect a
  closed set of statically shaped operators for fusion and
  generate code for CPU/GPU.

\relay's fusion algorithm addresses weaknesses of previous approaches by representing
  \textit{all} operators in a secondary IR.
\relay operators are backed by a TVM compute expression that
  describes operations in a high-level DSL that resembles Einstein notation
  but omits low-level scheduling details.
TVM's separation of compute and scheduling provides many favorable qualities
  for \relay's fusion algorithm.
It enables producing shape-specialized fused operators for an open set of operators,
  fusing arbitrary-length chains of operators (not just pairwise combinations),
  and handling operators with multiple outputs and nonlinear consumer-producer patterns.
TVM is also able to reschedule after fusion and perform further optimization via auto-tuning.
\relay performs fusion in two steps, detailed below.

\subsubsection*{Extraction}

First, \relay identifies subexpressions containing
  fusion-eligible and factors them into local functions that
  are marked as primitive.
Primitive functions can later be lowered to platform-specific
  code.
Fusion-eligible subexpressions are identified by constructing a
  directed acyclic graph (DAG) representing data flow between operators.
As the dataflow DAG is acyclic, it allows for the simple construction
  of a post-dominator tree.
Subexpressions are grouped into equivalence classes
  determined by their immediate post-dominator.
The use of the post-dominator tree enables fusion
  between non-linear producer-consumer relationships;
  for example, \relay can fuse diamond-shaped data-flow relations,
  where an input is used by multiple parallel operator chains
  that are combined again by a later operator.
Finally, \relay constructs an expression from each equivalence class,
  collects the expressions' free variables,
  constructs a function with the expression as the body and the free variables
  as parameters,
  and marks it as primitive.

\subsubsection*{Lowering}

In a second step, the \relay compiler converts the generated primitive
  function into platform and shape specific code.
For each operator, \relay collects the high-level \tvm expression that represents it,
  then combines them into an aggregate expression that represents the fused operation.
Generating code using \tvm also requires producing a schedule.
It is possible to use \tvm's default schedule to generate code for a single operation,
  but the default schedule does not support fusion.
In order to generate code for the combined expression, we must generate a
  master schedule based on the set of operations being fused.
The fusion algorithm analyzes the expressions to select a master
  schedule, the master schedule will perform the appropriate scheduling
  actions to generate fused code, such as inlining loops, or reorganizing
  computation.
By combining the master schedule with the fused computation,
  \relay is able to produce an optimized version of the operator
  for any platform supported by TVM.
For example, a related project by one of the co-authors implemented
  a RISC-V backend which immediately obtained full operator fusion
  with no new code.
Due to the \relay compiler's integration with AutoTVM, we can further
  optimize fused operations by performing auto-tuning on the master
  schedule template to obtain the best performance.

\subsection{Quantization Framework}
\label{sec:quant}

\begin{figure*}[htbp!]
  \centering
  \includegraphics[scale=0.53]{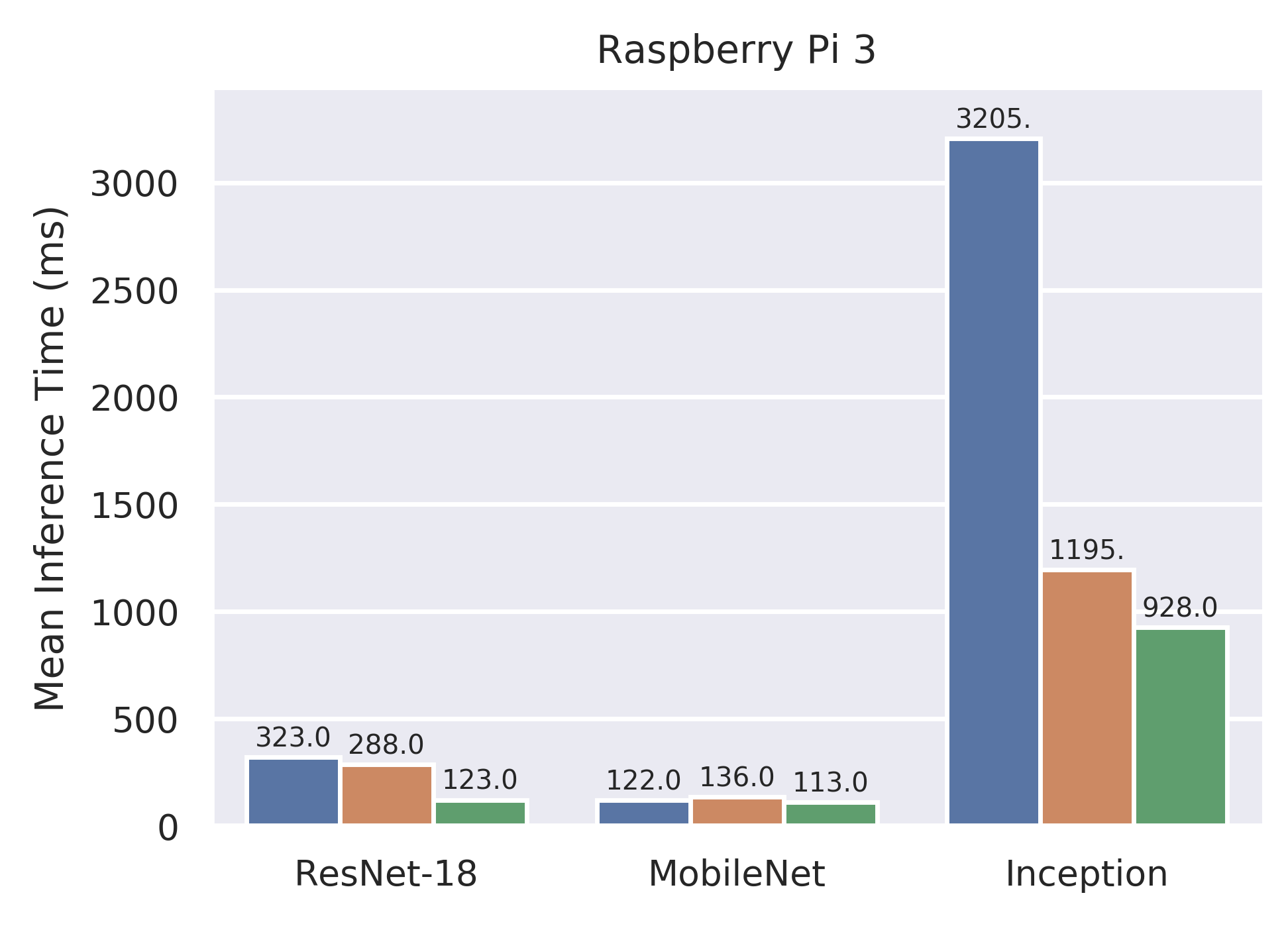}
  \hspace{-0.075in}
  \includegraphics[scale=0.53]{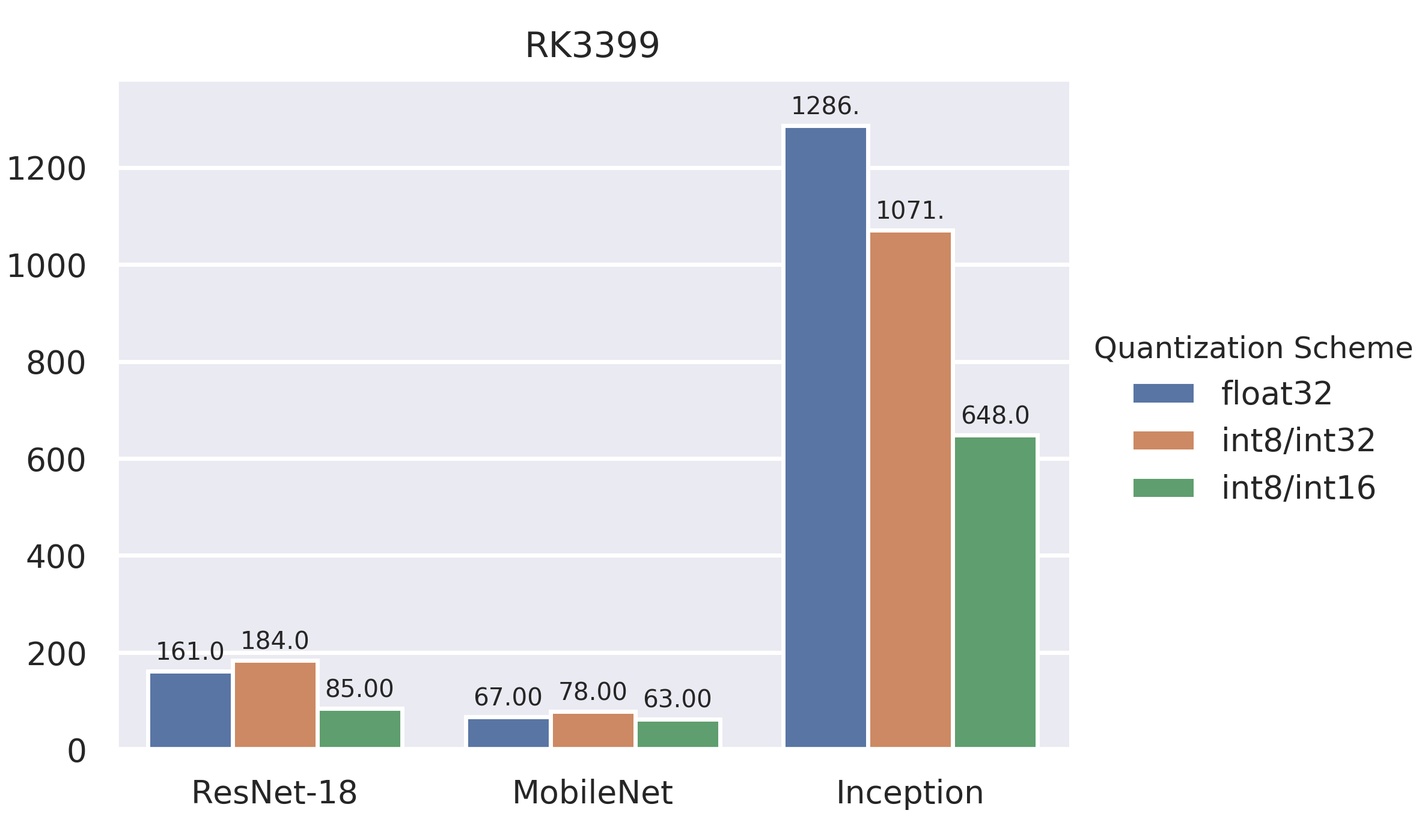}
  \includegraphics[scale=0.53]{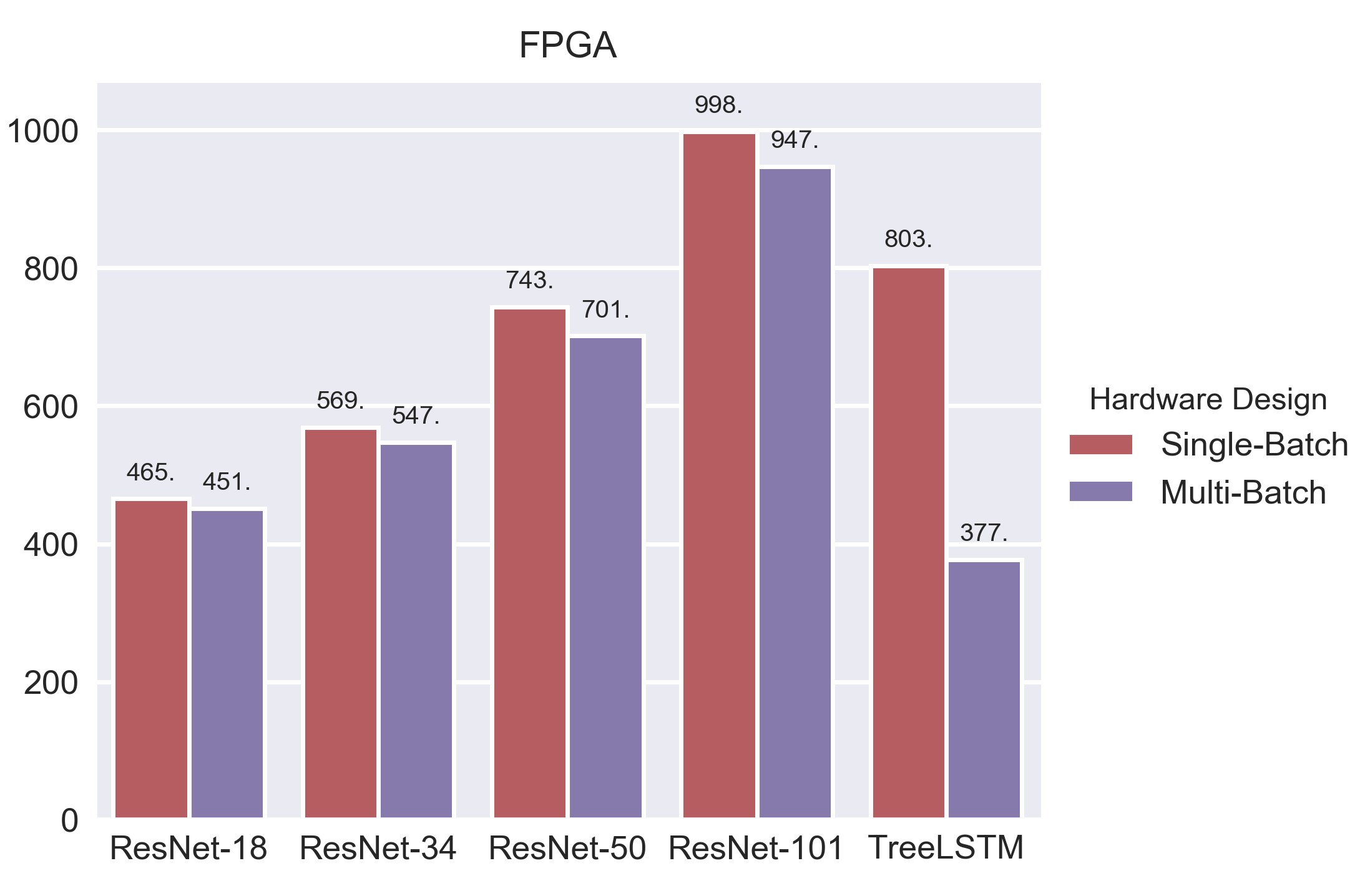}
  \caption{\textmd{
    \textit{(left)} Inference time of vision DNNs on low-power platforms using
      different data types.
    \relay allows us to reduce inference time on power-constrained devices by
      easily substituting \texttt{float32} multiplications with \texttt{int8}
      multiplications and \texttt{int16} or \texttt{int32} accumulations (denoted
      as \texttt{int8}/\texttt{int16} and \texttt{int8}/\texttt{int32}, respectively).
    We used 1000 trials for each model.
    \textit{(right)}
    Batch-size-normalized inference time of vision DNNs and a TreeLSTM running on two DNN accelerator variants implemented on an edge FPGA.
    One accelerator performs single-batch inference, while the other implements multi-batch inference.
    The two hardware designs have the same number of compute units that are arranged differently to take advantage of different types of tensor computation.
    \relay applies a multitude of graph-level transformations required to run different workloads onto these DNN hardware designs.
    We used 12 trials for each model.
  }}
  \label{fig:portability-eval}
\end{figure*}

Deep learning is constrained by memory, compute, and accuracy.
Accuracy is often the only metric optimized by machine learning
  researchers, leading to compute- and memory-hungry models.
The sheer number of parameters and the requisite compute
  makes deploying models to resource-limited devices,
  such as in mobile or IoT, challenging.
Even in non-edge devices, the compute cost of using
  datatypes like FP32 is large and computing with mixed precision
  or reduced precision can aid performance.
Unfortunately, reducing bit-width is not a silver bullet and
  can dramatically harm model accuracy.
The tradeoffs between these quantities has lead to the study of quantized neural networks,
  the process by which NNs are modified to use a smaller precision
  or non-standard datatypes to improve throughput and memory usage.
Quantization is particularly essential for supporting many accelerators due to
  their restricted set of datatypes.

State-of-the-art work on quantization demonstrates a number of tradeoffs
  between different quantization techniques,
  with the best often determined by platform and model type~\cite{krishnamoorthi18}.
Most DL frameworks have chosen a specific set of fixed quantization
  schemes and datatypes due to the effort required to manually implement operators for
  each pairing.

Instead, \relay includes a generic, compiler-based quantization flow that supports a diverse set
  of quantization schemes and can automatically generate code for each one.
\relay provides a general-purpose program-rewriting framework that can be extended
  with per-operator rules, which can annotate inputs and outputs with a datatype
  and precision to quantize to.
Users can overload \relay's existing quantization rewriting rules or add new ones
  to implement different quantization strategies, enabling users to choose between
  signed or unsigned integers or different rounding strategies, such as
  floor, ceiling, or stochastic rounding.

\begin{figure}[h]
  \includegraphics[height=5cm]{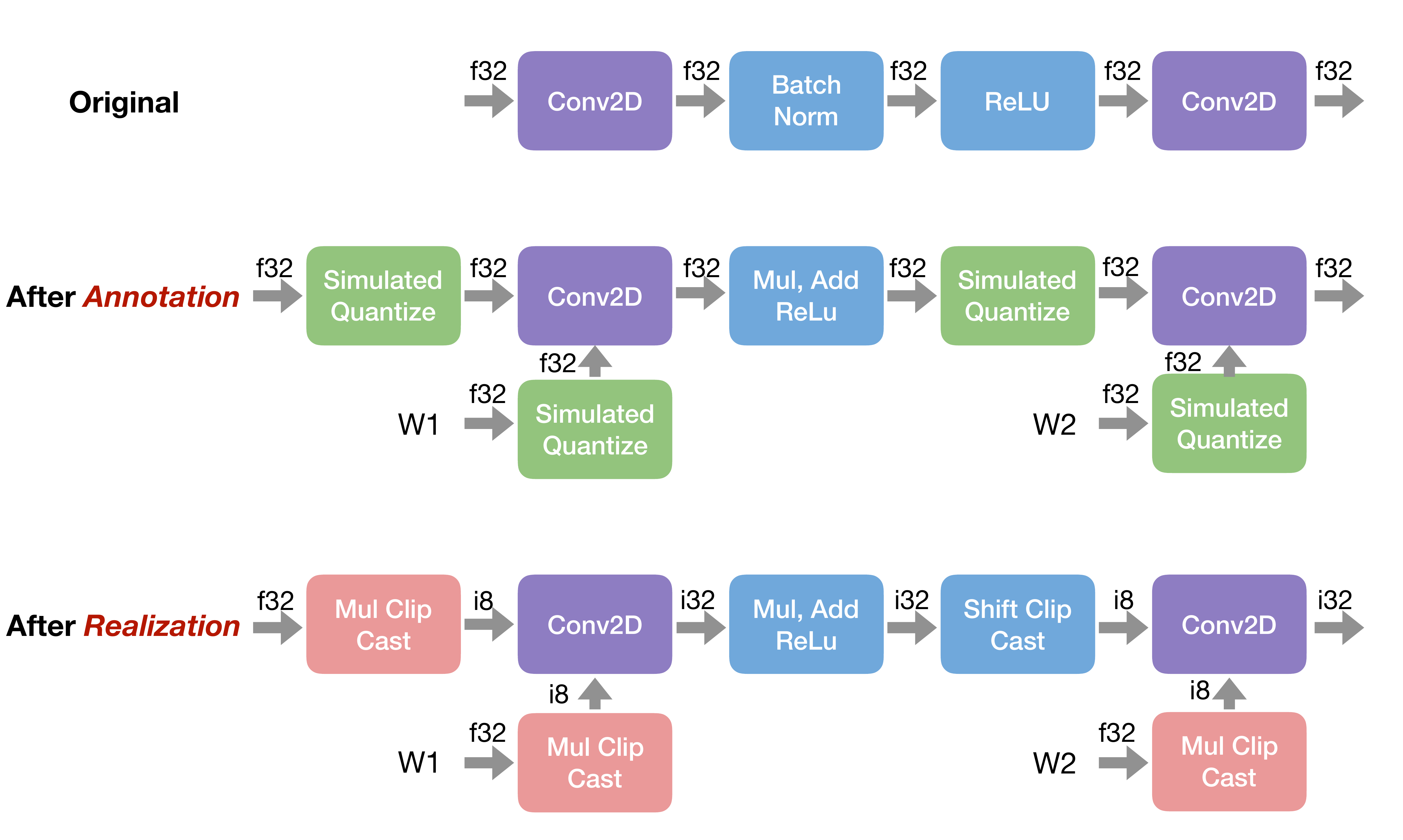}
  \caption{\textmd{The top graph represents the dataflow graph of operators after annotation,
  and the bottom graph represents the result of quantization.}}
  \label{fig:quant_flow}
\end{figure}

Figure \ref{fig:quant_flow} illustrates the rewriting process.
Furthermore quantization is expanded to standard
  \relay operators, which perform the scaling.
Due to this choice, \relay can then fuse these elementwise operations
  into the original operator, resulting in a brand-new quantized operation.
Finally, \relay can subsequently apply further optimizations like
  layout transformation, accelerator-specific packing, or
  auto-tuning to further improve performance or portability.
This enables the generation of customized quantized operators
  for user-provided schemes and operators,
  not limiting users to a single scheme.

We will now detail the three steps of the generic quantization flow:
  annotation, calibration, and realization.

\subsubsection*{Annotate}

Annotation rewrites the program to insert simulated quantization operations
  according to annotation rule for each operator.
Each input or output to be quantized is passed to \texttt{sim$Q$},
  an operator that simulates the effect of quantization (for example, from a 32-bit
  floating point value to an 8-bit integer value).
\texttt{sim$Q$} has a set of parameters that must then be calibrated in order to
  correctly quantize the graph, namely the bits, the scale, and the range.
\texttt{sim$Q$} simulates quantization on
  the unquantized type; that is, it performs computation on the unquantized type
  and then scales it to the target type.
By computing on the unquantized type, \relay can later calibrate the parameters to
  \texttt{sim$Q$} a necessary step to preserve accuracy of the model.
\vspace{-0.025in}
\[
  \texttt{sim$Q$}\left(x, \beta, \sigma, \rho\right) = \dfrac{\texttt{clip}\left(\texttt{round}\left(x / \rho \cdot 2^{\beta - \sigma}\right)\right) \cdot \rho}{2^{\beta - \sigma}}
\]
\subsubsection*{Calibrate}
As seen above \texttt{sim$Q$} has an input $x$, as well as a number of parameters
  $\beta$, $\sigma$, and $\rho$.
\texttt{sim$Q$}'s parameters control the mapping between the quantized and unquantized type
  and must be calibrated, without calibration the model can be wildly inaccurate.
We must perform an auxiliary optimization task to find the appropriate
  setting for these parameters.
The \relay compiler supports a variety of strategies for setting these
  parameters.
The first strategy implemented is a hyper parameter sweep of a
  single global scale until such a scale is found that does not result
  in overflow.
Another approach is a vision specific scheme which uses
  a per-channel scale, and optimizes the scales using a
  simple mean-squared error loss.
Finally an approach adopted from MxNet uses a
  KL-divergence based loss to optimize the
  quantization scales.

\subsubsection*{Realize}

Finally, after the algorithm has set the parameters appropriately,
  it applies realization,
  which transforms the \texttt{sim$Q$} operator into the below
  quantization operator.
\vspace{-0.05in}
\[
  Q\left(x, \rho, \beta, \sigma\right) = \texttt{cast}\left(\texttt{clip}\left(\texttt{round}\left(x / \rho \cdot 2^{\beta-\sigma}\right), \texttt{qtype}\right)\right)
\]
The produced operator performs the necessary scaling
  by realizing the operation as a sequence of finer-grained
  operators such as multiplication and rounding.
The output of original operator is now immediately scaled
  by this new operation.
Due to \relay's handling of fusion
  we are able fuse these scaling operations directly into
  to the original operator, transforming a convolution
  from \verb|fp32| to a type such as \verb|int4|.

\subsection{Partial Evaluator}
\label{sec:partial_eval}
Existing deep learning IRs have relied on
  a mixture of staging and constant evaluation
  in order to optimize user programs.
Partial evaluation is a generalized form of constant
  evaluation that can reduce partially constant
  programs.
A partial evaluator (PE) allows the use of high-level abstractions
  without limiting code that \textit{could} in practice be
  compiled to a particular target.
\relay is the first compiler to apply partial evaluation
  techniques to deep learning, the
  core approach of which is based on \cite{pe_ref}.
Partial evaluation, when composed with other
  optimizations like fusion, yields a variety
  of useful optimizations without requiring
  a separate implementation of each.
For example, the partial evaluator can be used to perform
  loop unrolling, which then enables further fusion,
  without any additional compiler passes.

\relay's partial evaluator works by defining a interpreter
  where the value domain is partially static values.
The partially static domain represents simple values,
  such as constant tensors, as themselves.
The representations
  of aggregate values mirror their structure; for example,
  tuples become a tuple of partially static values.
The partially static domain represents dynamic values,
  which may not be known until execution time,
  alongside the static values traditionally supported by
  constant evaluators.
Our partial evaluator must solve two important problems:
  managing effectful computations and handling references.
In order to handle effects, the evaluator keeps the generated
  program in A-normal form to ensure effects are properly ordered
  and restrict duplication of effectful computations.
The partial evaluator supports references by
  simulating the store at partial evaluation time.
The explicit store is threaded throughout execution
  and provides a flow-sensitive PE.
Finally, the evaluator constructs a new program with
  the static subcomputations evaluated away.

\subsection{Accelerator-Specific Optimizations}
\label{sec:accel-opts}


This subsection focuses on a subset of optimizations necessary to
  compile \relay to deep learning hardware accelerators.
Although DL accelerators form a diverse family of designs,
  one property they have in common is a restricted computing model.
This means that some individual accelerators
  may not be able to solely execute many \relay programs.
For example, many accelerators cannot execute unbounded loops,
  requiring some computation to be scheduled on a host device
  like the CPU.

\textit{Axis scale folding} is an optimization that removes scaling
  operations that occur before or after convolution-like operators.
The multiplication by a scalar is moved through a convolution towards
  its constant inputs, such as parameters.
By moving the scaling operation to a constant weight, we are able
  to compute away the scale using the partial evaluator.
This optimization is required for certain accelerators that lack scalar multipliers~\cite{moreau2018vta}.
In order to target these accelerators,
  we must eliminate \textit{all} scalar operations.

\textit{Parallel convolution combination} is a specialized
  optimization that fuses multiple 2D convolutions that share the same input.
The goal of this pass is to produce a larger kernel for the GPU,
  as each kernel launch on the GPU has overhead.
It was designed with the Inception network \cite{inception} in mind, as it
  contains blocks of convolutions that share the same input.
The entire parallel convolution combination pass,
  including documentation and tests,
  required fewer than 350 lines of code and was contributed
  by a non-\relay affiliated undergraduate student
  in their first contribution to our codebase.

\section{Evaluation}
\label{sec:eval}

%
%
%

\begin{figure*}[htp!]
  \centering
  \includegraphics[scale=0.53]{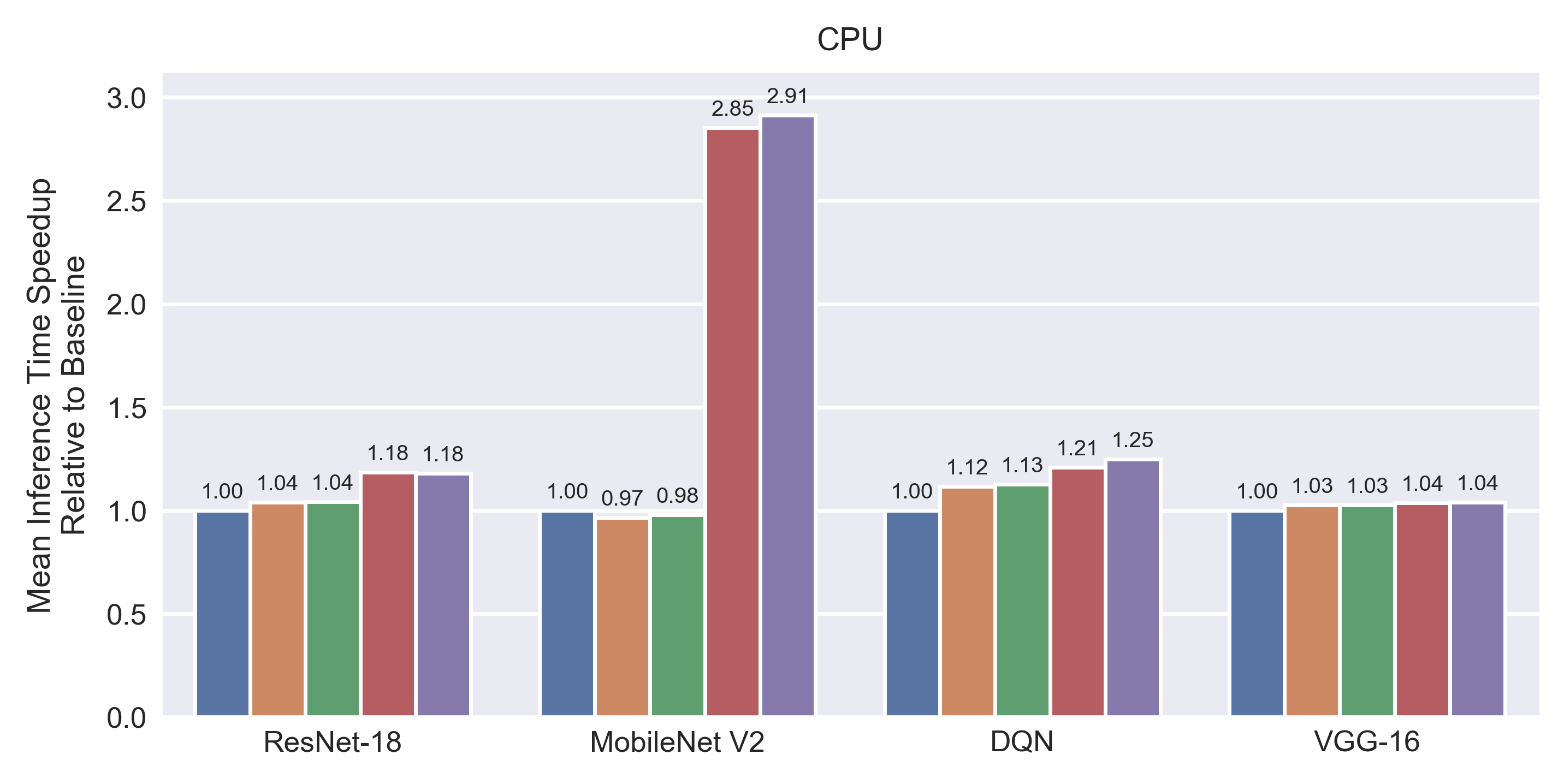}
  \includegraphics[scale=0.53]{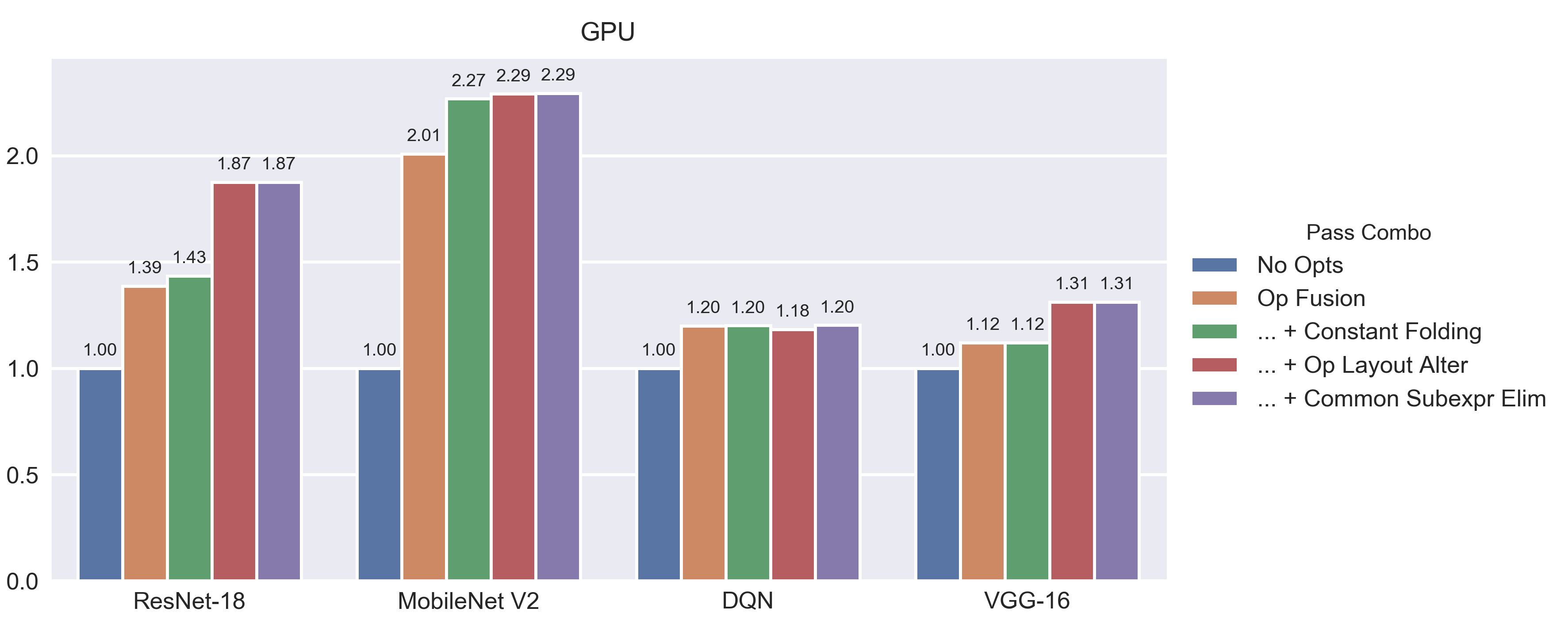}
  \caption{\textmd{
    Speedup from successively layering compiler passes in \relay on CPU
    (AMD Ryzen Threadripper 1950X) and GPU (Nvidia Titan-V),
      relative to no optimizations at all.
    The ``Op Fusion'' bars represent the application of operator fusion,
      the ``... + Constant Folding'' bars represent the application of operator fusion \textit{and} constant folding,
      and so on.
    The full list of passes used is as follows:
      \textit{operator fusion};
      \textit{constant folding};
      \textit{operator layout alteration}, which transforms the data layouts of operators for better cache performance;
      and \textit{common subexpression elimination}.
    We find that composing passes can steadily increase performance.
    The effectiveness of each pass is both model- and device-dependent.
    In particular,
      the most effective passes for CPU and GPU are operator layout alteration and operator fusion,
      respectively.
  }}
  \label{fig:composability-eval}
\end{figure*}

We evaluate \relay's ability to provide expressivity, composability, and portability,
  without compromising on performance.
In particular, our evaluation is composed of three parts:
\begin{enumerate}
  \item \textbf{\relay expresses diverse workloads}: Despite increasing
    expressiveness, \relay's performance is competitive with the
    state of the art on popular models.
  \item \textbf{\relay enables composable optimizations}: \relay
    supports composing program transformations to incrementally improve performance.
  \item \textbf{\relay handles challenging backends}: \relay can compile
    models to execute efficiently on a variety of
    backends, such as FPGA accelerators, which require quantization, layout
    optimizations, and bit-packing transformations.
\end{enumerate}

We evaluated the following vision models:
  \textit{Deep Q-Network (DQN)}, a CNN that achieved state-of-the-art performance
  on 49 Atari games in 2015;
  \textit{MobileNet}, a CNN designed for image recognition on mobile and
  embedded devices;
  \textit{ResNet-18}, a CNN for image recognition that achieved state-of-the-art
  performance on ImageNet detection tasks in 2015;
  and \textit{VGG-16} (named for the Visual Geometry Group at Oxford),
    a CNN for image recognition that achieved top-2 performance in the 2014 ImageNet Challenge.
  \cite{dqn, mobilenet, resnet, vgg}.
We evaluated the following NLP models:
  \textit{CharRNN}, a generator character-level
  RNN from a PyTorch tutorial;
  \textit{TreeLSTM}, a generalization of LSTMs to
  tree-structured network topologies;
  and \textit{RNN}, \textit{GRU}, and \textit{LSTM}, a selection of models from the Gluon
  Model Zoo
  \cite{pytorch_rnn_tut, tree_lstm, gluon_model_zoo}.

\subsection{Experimental Methodology}
Because we only evaluate inference in this paper,
  we frequently make use of random inputs to models when measuring
  performance.
There were two exceptions where we evaluated on real data because
  it was readily available: CharRNN and TreeLSTM.
For each experiment,
  we run 10 untimed ``warm-up'' iterations to ensure any effects from caching and
  JIT compilation are excluded from the timed runs.
The vision and NLP experiments (from Section~\ref{sec:eval_opts} and Section~\ref{sec:perf-gpu})
  were run on a machine with an AMD Ryzen Threadripper 1950X 16-Core CPU,
  an Nvidia Titan-V GPU,
  and 64 GB of RAM.
For the vision workloads,
  we used TVM's graph runtime as the executor,
  and for the NLP workloads,
  we used \relay's AoT compiler.
The low-power vision experiments from Section~\ref{sec:low-power} were run on multiple edge-class ARM development boards: a Raspberry Pi 3, a Firefly RK3399, and an Pynq-Z1 FPGA platform.
We implement our DNN accelerators on a Zynq-7020 low-cost FPGA, and clock them at 100MHz.
We used the following software versions:
  CUDA version 10.0,
  CuDNN version 7.5.0,
  TVM commit \texttt{e518fe1c}\footnote{NLP experiments required extensions to the MxNet importer that will be made public later},
  MxNet version 1.5.0,
  PyTorch version 1.2.0,
  and TensorFlow version 1.14.0.

\subsection{\relay Expresses Diverse Workloads}
\label{sec:perf-gpu}

\begin{figure*}[htbp!]
  \centering
  \includegraphics[scale=0.68]{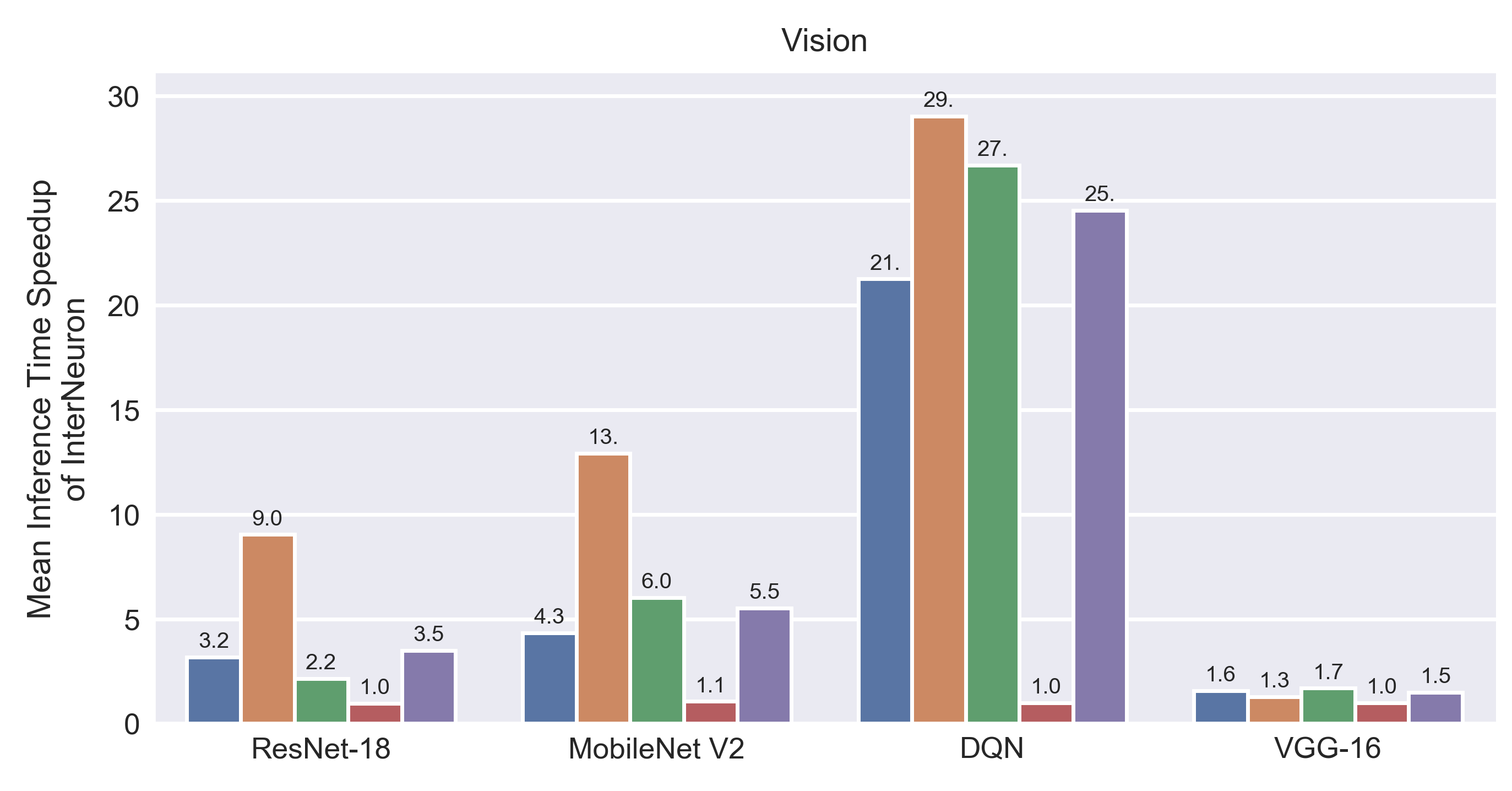}
  \includegraphics[scale=0.68]{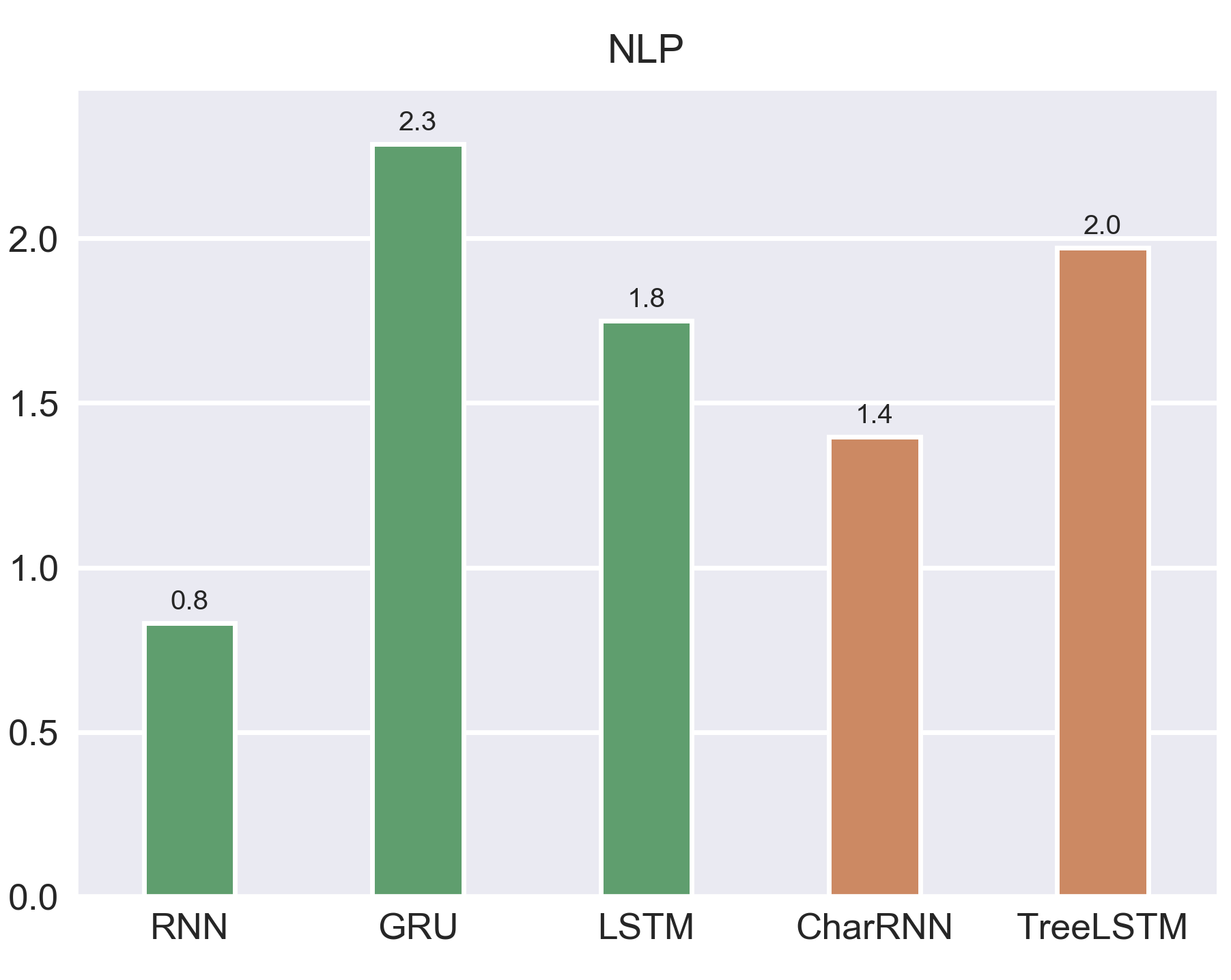}
  \includegraphics[scale=0.68]{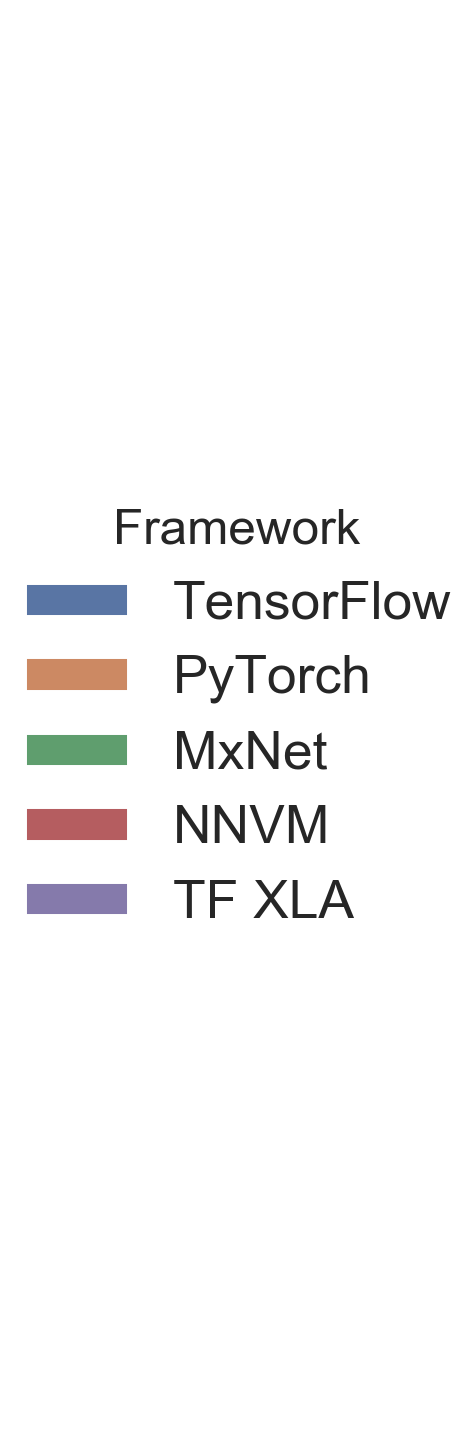}
  \caption{\textmd{
    Inference speedup of \relay relative to popular frameworks
      on vision and NLP benchmarks.
    The vision benchmarks used an NVIDIA Titan-V GPU, and the NLP benchmarks ran on CPU only.
    We ran 1000 trials for each model, except for CharRNN, on which we used 100 trials.
    \relay matches the performance of NNVM on vision but additionally supports NLP,
      where \relay provides performance competitive to the state of the art (up to
        2.3$\times$ speedup over MxNet on GRU).
  }}
  \label{fig:expressivity-eval}
\end{figure*}

An age-old story in compilers literature is that increasing expressivity
  impacts the global performance of the system.
We set out to build zero-cost abstractions for \relay,
  governed by Stroustrup's principle, ``What you don't use, you don't pay
  for'' \cite{bjarne}.
We demonstrate that we achieve competitive performance on a wide set of CNNs that are well supported by existing frameworks.
We evaluated inference time for two classes of workloads: computer vision and natural language processing.
We compared \relay to \nnvm,
  TensorFlow, TensorFlow-XLA (Accelerated Linear Algebra), PyTorch, and MxNet.
Results are summarized in Figure~\ref{fig:expressivity-eval}.

\subsubsection*{Vision Evaluation}

We ran each model with
  batch size 1, a common setting in inference tasks.
\relay achieves performance on par with \nnvm
  and outperforms TensorFlow, TensorFlow-XLA, MxNet, and
  PyTorch on every benchmark.
\relay's ability to apply aggressive inter-operator optimizations
  enables it to outperform existing frameworks.
Operator fusion over long chains of operations is particularly effective,
  because it can generate \textit{new} hardware-specific fused implementations.

\subsubsection*{NLP Evaluation}
Implementations of the NLP models were not available in all frameworks;
  we used MxNet baselines for RNN, GRU, and LSTM and PyTorch for CharRNN and TreeLSTM.
NLP workloads feature control flow,
  which makes them more challenging to optimize.
\relay performs better than MxNet on GRU and LSTM
  because they are implemented in Python using
  MxNet's looping constructs.
However,
  MxNet outperforms \relay on the Gluon RNN,
  because it uses a hardcoded optimization to unroll the RNN,
  whereas \relay expresses it as a loop without any unrolling.
PyTorch instead uses handwritten and heavily optimized
  C implementations of the recursive network cells.
Despite this,
  our pure \relay implementation outperforms PyTorch by 1.4$\times$ on
  CharRNN and 2$\times$ on TreeLSTM.
This speedup comes from \relay's ability to compile \textit{entire}
  models with complex control flow (e.g., CharRNN) to a single lean binary.

\subsection{\relay Enables Composable Optimizations}
\label{sec:eval_opts}

We demonstrate that \relay facilitates composable optimizations
  by evaluating vision workloads under both general-purpose and DL-specific compiler passes.
Figure~\ref{fig:composability-eval} shows mean inference speedup relative to
  no optimizations as \relay applies optimizations more aggressively.
We find that performance gains vary significantly between each device-model pairing.
Most networks benefit greatly from operator layout alteration on CPU and operator fusion on GPU.
On both CPU and GPU,
  VGG-16 is resistant to most optimizations,
  because the architecture primarily consists of back-to-back convolutions,
  which are not fusable.
Elementwise operations \textit{can} be fused,
  so we see greater improvements on ResNet and MobileNet,
  which both feature elementwise adds from residual connections.
It's unsurprising that MobileNet fares so well on CPU---it is \textit{designed} to run well on CPU.
Nature-DQN has simple operators,
  which don't benefit from layout alterations,
  whereas ResNet-18 and VGG-16 are dense convolutional neural networks,
  which \textit{do} benefit from layout transformations.
Overall, these results show that \relay lets us compose optimizations
  in a way that is beneficial to diverse workloads.

\subsection{\relay Handles Challenging Backends}
\label{sec:low-power}
To demonstrate portability,
  we evaluate two sets of optimizations:
    those that are merely \textit{beneficial} for low-power platforms and
    those that are \textit{necessary} to target hardware accelerators.

\subsubsection*{Quantized Inference on ARM Platforms}
To demonstrate the effectiveness of our generic quantization (see Section~\ref{sec:quant}),
  we use \relay to evaluate both \textit{accuracy} and \textit{performance} of different
  quantization schemes on vision workloads.
To evaluate \textit{accuracy},
  we tested various quantization schemes
  (denoted $m/n$ for $m$-bit quantization and $n$-bit accumulation)
  against a \texttt{float32} baseline on three vision models,
  as shown in the table below:
\begin{center}
  \begin{tabular}{|c|c||c|c||c|c|}
    \hline
    \multicolumn{2}{|c}{\textbf{ResNet-18}} & \multicolumn{2}{c}{\textbf{MobileNet V2}} & \multicolumn{2}{c|}{\textbf{Inception V3}} \\
    \multicolumn{1}{|c}{QS}    & \multicolumn{1}{c}{Acc.}   &  \multicolumn{1}{c}{QS}  & \multicolumn{1}{c}{Acc.}  & \multicolumn{1}{c}{QS}  & \multicolumn{1}{c|}{Acc.} \\
    \hline
    \texttt{fp32} & 70.7 \%    & \texttt{fp32} & 70.9 \%       & \texttt{fp32} & 76.6 \% \\
    8/32         & 69.4 \%    & 8/32         & 66.9 \%       & 16/32        & 76.6 \% \\
    8/32         & 69.4 \%    & 8/16         & 66.9 \%       & 8/32         & 75.2 \% \\
    \hline
  \end{tabular}
\end{center}
Figure~\ref{fig:portability-eval} shows the results of different
  levels of quantization on \textit{performance} when applied to the Raspberry Pi 3
  and Firefly RK3399 ARM-based platforms.
The numbers show that as we opt for a more aggressive quantization scheme
  (e.g., 8/16),
  we achieve much improved performance with hardly a drop in accuracy.
Interestingly,
  on some model/platform pairs,
  the \texttt{int8/int32} scheme performs slightly worse than \texttt{float32} on both platforms,
  which likely stems from the existence of faster hardware intrinsics for 16-bit operations on these systems.

\subsubsection*{Targeting Deep Learning Accelerators on FPGAs}
We demonstrate that \relay can support specialized hardware by compiling vision and
NLP workloads onto two DNN accelerator designs (\texttt{single-batch}, \texttt{multi-batch}) we generate in-house.
%
%
%
The two DNN designs have the same number of MAC (multiply and accumulate) units which are
arranged differently to expose different compute intrinsics to the compiler.
%
%

We evaluate batch-size-normalized inference time on the accelerator designs
  on a mix of vision and NLP workloads:
  ResNet-18, ResNet-34, ResNet-50, ResNet-101 \cite{resnet};
  and TreeLSTM in Figure~\ref{fig:portability-eval}.
The ResNets have high arithmetic intensity
due to 2D convolutions, while the NLP workload is memory bound due to the lower-intensity vector-matrix
multiplication used in the LSTM cells.

We show that running the workloads
  on the \texttt{multi-batch} DNN accelerator improves throughput on all workloads at the cost
  of naturally increasing inference latency.
Batching is not compelling on ResNet because it increases the arithmetic
  intensity of a workload that is already compute-bound.
On the other hand, TreeLSTM presents a compelling target for batching due to the
  memory bound nature of the LSTM cell computation.
%
While these two accelerator variants have the same peak throughput on paper, we show that \relay let us
  evaluate more nuanced end-to-end performance numbers across different workloads.


These experiments demonstrate \relay's ability to target current and future deep learning architectures,
and make informed decision on what hardware design to choose across different workloads.

\section{Conclusion}
\label{sec:conclusion}

This paper introduced \relay,
  a high-level IR that enables end-to-end optimization of
  deep learning models for a variety of devices.
In particular, \relay provides a design for \textit{extensibility}.
In addition to representing, optimizing, and executing models defined in
  popular frameworks, we use \relay's design to
  define a combination of traditional and domain-specific optimizations.
\relay's approach can be adopted by other DL frameworks to
  implement IRs that can support extension \textit{without} compromising
  on \textit{performance}, \textit{expressivity}, \textit{composability}, or \textit{portability}.
With its extensible design and expressive language, \relay serves
  as a foundation for future work in applying compiler
  techniques to the domain of deep learning systems.


\bibliographystyle{plain}
\bibliography{paper}

\end{document}